\documentclass[11pt,a4paper]{article}
\usepackage[hyperref]{emnlp2020}
\usepackage{times}
\usepackage{latexsym}
\usepackage{url}
\usepackage{amsmath}
\usepackage{enumitem}
\setlist{nosep} 
\usepackage{CJKutf8}

\usepackage{tikz}
\usepackage{pgfplots}
\pgfplotsset{width=7.5cm,compat=1.12}
\usepgfplotslibrary{fillbetween}

\usepackage{graphicx}
\graphicspath{ {./images/} }

\newcommand{\zh}[1]{\begin{CJK*}{UTF8}{gbsn}#1\end{CJK*}}

\aclfinalcopy 

\title{Learning to Pronounce Chinese \\ Without a Pronunciation Dictionary}

\author{Christopher Chu, Scot Fang, and Kevin Knight  \\
DiDi Labs \\
4640 Admiralty Way \\
Marina del Rey, CA 90292 \\
{\tt \{chrischu,scotfang,kevinknight\}@didiglobal.com}}

\date{}

\begin{document}
\maketitle

\begin{abstract}
We demonstrate a program that learns to pronounce Chinese text in Mandarin, without a pronunciation dictionary.  From non-parallel streams of Chinese characters and Chinese pinyin syllables, it establishes a many-to-many mapping between characters and pronunciations.  Using unsupervised methods, the program effectively deciphers writing into speech.  Its token-level character-to-syllable accuracy is 89\%, which significantly exceeds the 22\% accuracy of prior work.
\end{abstract}


\section{Unsupervised Text-to-Pronunciation}

Many papers address the construction of automatic grapheme-to-phoneme systems using rules or supervised learning, e.g. \cite{Berndt1987EmpiricallyDP,zhang02,xu04,Bisani2008JointsequenceMF,Peters2017MassivelyMN}.

The task of {\em unsupervised} grapheme-to-phoneme conversion is introduced by \citet{Knight99acomputational}.  Given two non-parallel streams:

\begin{itemize}
    \item A corpus of written language (characters).
    \item A corpus of spoken language (sounds).
\end{itemize}

\noindent
the goal is to build:

\begin{itemize}
    \item A mapping table between the character domain and the sound domain.
    \item A proposed pronunciation of the written character sequences.
\end{itemize}

\noindent
Motivated by archaeological decipherment, \citet{Knight99acomputational} view character sequences as ``enciphered'' phoneme sequences.  Their evaluation compares the proposed pronunciations with actual pronunciations.  With a noisy-channel expectation-maximization method, they obtain 96\% phoneme accuracy on Spanish, 99\% on Japanese kana, but only 22\% syllable accuracy on Mandarin Chinese.

\begin{figure}
\centering
\includegraphics[scale=0.6]{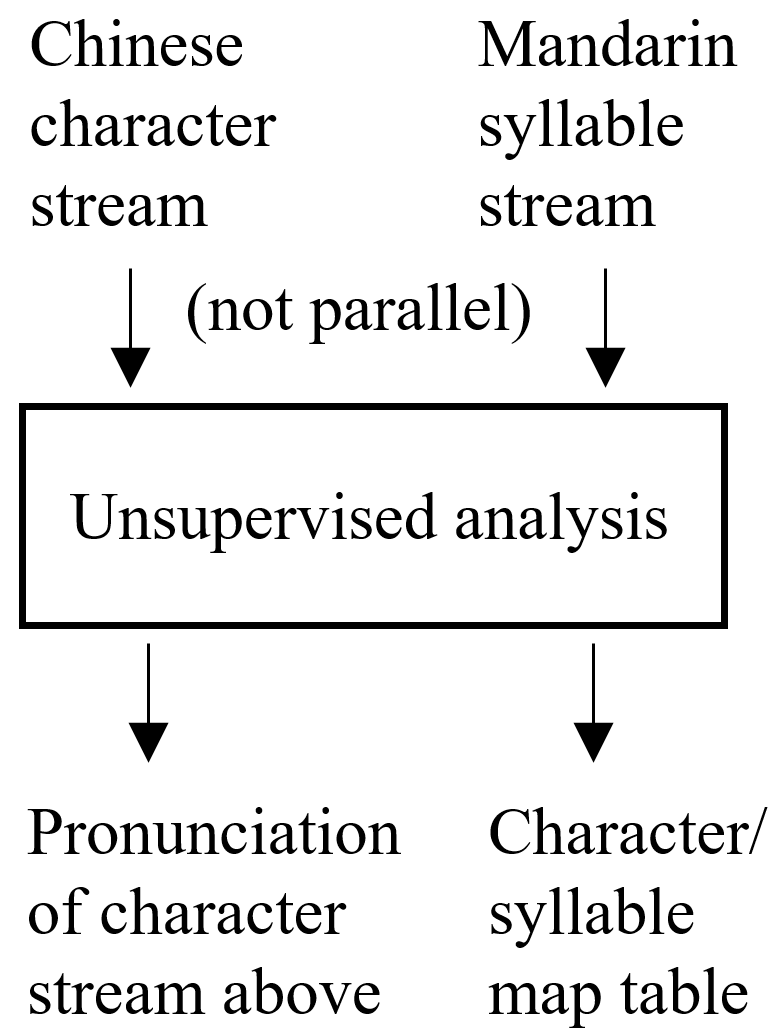}
\caption{Learning to pronounce Chinese without a dictionary.}
\label{overview}
\end{figure}

In this paper, we re-visit the task of deciphering Chinese text into standard Mandarin pronunciations (Figure~\ref{overview}).  We obtain an improved 89\% syllable accuracy.  We further explore exposing the internals of characters and syllables to the analyzer, as Chinese characters sharing written components often sound similar.

We find it compelling that pronunciation dictionaries are largely redundant with non-parallel text and speech corpora, even for writing systems as complex as Chinese.  We also expect results may be of use in dealing with novel ways to write Chinese, such as N\"ushu script \cite{Zhang2016FromIT}, with acoustic modeling of other Chinese languages and dialects, and with novel ways to phonetically encode and decode Chinese in online censorship applications \cite{Zhang2014BeAA}.

\section{Chinese Writing}

The most-popular modern Chinese writing system renders each spoken syllable token with a single character token ({\em hanzi}).  There are over 400 syllable types in Mandarin\footnote{In this paper, we use standard {\em pinyin} syllable representation, and we refer strictly to Mandarin pronunciation.} and several thousand character types.  The mapping is many-to-many:

\begin{itemize}
    \item Almost every syllable type can be written with different characters (eg, {\em zhong} $\rightarrow$ \zh{\{中, 重, 肿, ...\}}).  The choice depends on context.  For example, the word {\em {\bf \em zhong}guo} (``China'') is written \zh{{\bf 中}国}, but {\em {\bf \em zhong}yao} (``important'') is written \zh{{\bf 重}要}.
    \item Only a few character types are heteronyms, whose pronunciation depends on context and meaning (e.g., \zh{了} $\rightarrow$ \{{\em le}, {\em liao}\}).
\end{itemize}

In addition, syllables have one of five tones, including the neutral tone (e.g., {\em d\=ang}, {\em d\'ang}, {\em d\v{a}ng}, {\em d\`ang}, {\em dang}).  Tones increase pronunciation ambiguity.  Most characters have single, unambiguous phonemic pronunciations (eg, \zh{中} $\rightarrow$ {\em zh\=ong}), but a portion can be pronounced with different tones depending on context (eg, \zh{{\bf 当}然}  $\rightarrow$ {\em {\bf \em d\=ang}r\'an}, but \zh{适{\bf 当}} $\rightarrow$ {\em sh\`{\i}}{\bf \em d\`ang}).

While most Chinese words have two syllables, individual characters carry rough semantic meanings (eg, \zh{中} = ``middle'', \zh{重} = ``weighty'').  So it is no accident that the same character is used to write semantically-similar words:

\begin{itemize}
    \item \zh{{\bf 中}国} (``China = middle kingdom''), \zh{{\bf 中}学} (``middle school''), \zh{市{\bf 中}心} (``city center'')
    \item \zh{{\bf 重}要} (``important''), \zh{{\bf 重}达} (``heavy''), \zh{{\bf 重}点} (``focus'')
    \item Similarly, the second syllable of ``website'' is spelled ``site'', not ``sight''.
\end{itemize}

Finally, many characters have loosely informative internal structure. For example, \zh{鸦} can be analyzed into two {\em character components}: \zh{牙} and \zh{鸟}.\footnote{Non-Chinese speakers may want to visually confirm that \zh{牙} and \zh{鸟}, suitably thinned and placed side-by-side, do indeed form the composite character \zh{鸦}.} 
Character components are sometimes a clue to pronunciation and/or meaning.  For example:
\begin{itemize}
\item The character \zh{鸦} (``crow'', {\em y\=a}) is composed of \zh{鸟} (meaning ``bird'') and \zh{牙} (sound {\em y\'a}).
\item The \zh{中} ({\em zh\=ong}) component of \zh{肿} (``swollen'') is a clue to its pronunciation {\em zh\v{o}ng}, though the \zh{月} (``moon'') component is more loosely suggestive of its meaning.
\item For a character like \zh{法} (``law''), the components ``water'' and ``go'' do not provide much of a phonetic or semantic clue.  This is the case with many characters.
\end{itemize}
\noindent
The vast majority of characters have two top-level components, arranged either side-by-side (as in the examples above), top-bottom, or outside-inside.  It should be noted that a top-level character component may often be recursively divided into further sub-components.

Generally speaking, it is impossible for a student to correctly guess the pronunciation or meaning of a new character, though their guess may be better than chance.

\section{Data Preparation}

From a Chinese Wikipedia dump,\footnote{\scriptsize linguatools.org/tools/corpora/wikipedia-monolingual-corpora} we remove all non-Chinese characters, then convert to simplified characters.  This forms our character corpus.

For our pronunciation corpus, we could record and transcribe Mandarin speech into {\em pinyin} syllables.  Instead, we simulate this.  We take a large subset of the Baidu Baike encyclopedia,\footnote{baike.baidu.com} but then immediately convert it to tone-marked {\em pinyin} syllables, by using a comprehensive dictionary\footnote{www.mdbg.net/chinese/dictionary?page=cc-cedict} of 116,524 words and phrases.  99.97\% of Baike character tokens are covered by this dictionary.

We substitute Baike character sequences with {\em pinyin} sequences in left-to-right, longest-match fashion.  This strategy works well most of the time.  For example, it correctly pronounces \zh{睡觉} as {\em shui jiao}, and \zh{觉得} as {\em jue de}, despite the ambiguity of \zh{觉}. However, it incorrectly pronounces \zh{想睡觉} because a dictionary entry \zh{想睡} matches the phrase before \zh{睡觉} = {\em shui jiao} can be applied; it also has trouble with single-character words like \zh{还}.




Using character sequences from Chinese Wikipedia and pinyin sequences from Baidu Baike is important.  If we alternatively divided Chinese Wikipedia into two parts, unsupervised analysis could easily exploit high-frequency boilerplate expressions like \zh{英語重定向：这是由英語名稱，指向中文名稱的重定向。它引導出英語標題至遵循命名常規的合適名稱，能夠協助編者寫作。} (``English Redirection: This is a redirect from the English name to the Chinese name. It guides the English title to a proper name that follows the naming convention and can assist the editor in writing.'')

We also {\em pinyin}-ize the first 100 lines (6059 characters) of our character corpus, as a gold-standard reference set, for later judging how well we phonetically decipher the character corpus. Unless stated otherwise, all results are for token accuracy on this reference set.

\begin{table}
    \centering
    \begin{small}
\begin{tabular}{|l|r|r|r|} \hline
& Tokens & Types & Singletons \\ \hline
Characters & 510m & 17,442 & 3,444 \\ 
(Wikipedia) &   &  & \\ \hline
Syllables & 264m & w/o tones: \hspace{1mm} 412 & 1  \\ 
(Baike)  &   &  w/ tones: 1506 & 5 \\ \hline
Test set & 6059 & 783 & 236 \\ 
(characters) & & & \\ \hline
\end{tabular} 
\end{small}
\caption{Token and type statistics for our non-parallel character and syllable corpora.  Singletons are one-count types.}
\label{data}
\end{table}

\begin{table*}
\begin{center}
\small
\begin{tabular}{|l|r|r|r|} \hline
& Exact {\em pinyin} match, with tone & Exact {\em pinyin} without tone & Partial {\em pinyin}  \\ \hline
Majority-class Baseline ({\em y\`u}) &  0.01  & 0.02 & 0.19 \\ \hline
Supervised Match 1  & 0.17  & 0.25 & 0.39 \\ \hline
Supervised Match 2  &  0.19 & 0.28 & 0.48 \\ \hline
\end{tabular}
\end{center}
\caption{Even with a partial pronunciation dictionary, it is difficult to predict exact pronunciation of a new written character from its components.  This table records accuracy of pronunciation guesses for characters 2001-3000 (by frequency), given pronunciations of characters 1-2000; for these types, {\em y\`u} is most frequent. Match~1 uses a character's second component, e.g., guessing (incorrectly) that \zh{耗} ({\em h\`ao}) is pronounced the same as \zh{毛} ({\em m\`ao}).  Match~2 uses either the first or second component, whichever is better.  Partial match credits either onset or rime, e.g., counting {\em h\`ao} for {\em m\`ao} as correct.}
\label{predict}
\end{table*}

Table~\ref{data} gives statistics on our corpora.  We release our data at {\em https://github.com/c2huc2hu/ unsupervised-chinese-pronunciation-data}.

We also record internal structure of syllables and characters.  For syllables, we separate onset and rime (eg, {\em zhang} $\rightarrow$ {\em zh} + {\em ang}).  

For characters, we employ the thorough graphical decompositions given in Wikimedia Commons,\footnote{commons.wikimedia.org/wiki/Commons:\\Chinese\_characters\_decomposition} which divide each character into (at most) two parts.  This allows us to find, for example, 44 characters that include the second component \zh{包} (\zh{咆}, \zh{孢}, \zh{狍}, \zh{炮}, etc).  We only use top-level decompositions in this work, forgoing any further recursive decompositions.

\section{Supervised Comparison Points}

Before turning to unsupervised methods, we briefly present two supervised comparison points.

First, if we had a large {\em parallel} stream of character tokens and their {\em pinyin} pronunciations, we could train a simple pronouncer that memorizes the most-frequent {\em pinyin} for each character type. Using the Baike data as processed above as a putative parallel resource, we obtain 99.1\% pronunciation accuracy on the test set.\footnote{Experimental results from here on out refer to no-tone pronunciation.}  The only errors involve ambiguous characters, showing that a deterministic character-to-pinyin mapping table---whether obtained by memorization or by unsupervised methods---is sufficient to solve the bulk of the problem.

Second, to investigate whether written character components predict pronunciation, we use gold pronunciations of the most common 2000 characters to predict pronunciations of the next 1000.  If a test character X has second (e.g., rightmost) written component Y, then we use the pronunciation of Y as a guess for the pronunciation of X.  We find this works 25\% of the time if we do not consider tones, and 17\% of the time if we do.  Table~\ref{predict} confirms that character components are only a loose guide to pronunciation, even with supervision.

Next we turn to unsupervised pronunciation, where we throw away pronunciation dictionaries and parallel data, working only from uncorrelated streams of characters and syllables.

\section{Unsupervised Vector Method}
\label{vector}

Borrowing from unsupervised machine translation, which learns mappings between words in different languages \cite{lample,artetxe18}, we attempt to learn a mapping between embeddings for characters and embeddings for {\em pinyin} symbols. We train fastText \cite{fasttext} vectors of dimension 300 and default settings on each of our corpora and use the MUSE system\footnote{https://github.com/facebookresearch/MUSE} to learn the relationship between the two vector spaces \cite{lample2018word}. 


There are two steps to this method: (1) map character vectors into {\em pinyin} space, (2) for each character type, find its nearest {\em pinyin} neighbor.  This gives us a table that maps character types onto {\em pinyin} types.  We apply this table to each character token of our 6059-character test set, obtaining token-level accuracy.

Unfortunately, this method does not work well.  Only 0.5\% of type mappings are correct, and token-level accuracy is similarly small.  Reversing the mapping direction ({\em pinyin} embeddings into character embedding space) does not improve accuracy.  It appears that the asymmetry of the mapping is difficult for the algorithm to capture.  Each {\em pinyin} syllable should, in reality, be the nearest neighbor of many different characters.  Moreover, the behavior of a {\em pinyin} syllable in running {\em pinyin} data may not be a good match for the behavior of any given character with that pronunciation.

Our next approach is to map words instead of characters. We break our long character sequence into a long word sequence, e.g., \zh{竞争 \hspace{1mm} 很 \hspace{1mm} 激烈} by applying the Jieba tokenizer\footnote{github.com/fxsjy/jieba} to Wikipedia.  We similarly break our long {\em pinyin} sequence into a long {\em pinyin}-word sequence, e.g., {\em wo xihuan chi jiaozi}, by applying the Stanford tokenizer\footnote{nlp.stanford.edu/software/segmenter.shtml} to pre-{\em pinyin}ized Baidu.  We build embeddings for types on both sides, and we again map them into a shared space. 

During the nearest-neighbor phase, we take each written-word and look for the closest {\em pinyin}-word, giving preference to {\em pinyin} words with the same number of syllables as the written-word.  If we cannot find a near neighbor with the correct number of syllables, we map to a sequence of {\em de}, the most common Chinese {\em pinyin} token.

\begin{table}
\centering
\small
\begin{tabular}{|l|l|l|r|} \hline 
Source & Target & Tone? & Accuracy \\ 
Character & Pinyin & no & 0.20\% \\ \hline
Character & Pinyin & yes & 0.05\% \\ \hline
Pinyin & Character & no & 0.15\% \\ \hline
Pinyin & Character & yes & 0.12\% \\ \hline
Character word & Pinyin word & yes & 81.41\% \\ \hline
\end{tabular}
\caption{Accuracy of vector-mapping approaches, measuring \% of character tokens we assign the correct (no tone) {\em pinyin} pronunciation to. Testing is on the first 6059 characters of the character corpus.}
\label{vecresults}
\end{table}

We find that matched word pairs are much more accurate than the individual character-{\em pinyin} mappings we previously obtained.\footnote{Similar to \citet{koehn20} and \citet{ney20}, we note that unsupervised translation techniques require certain types of data to work well.}  To get token-level pronunciation accuracy, we segment our 6059-token character test set, apply our learned mapping table, and count how many characters are pronounced correctly.  Table~\ref{vecresults} shows that the accuracy of this method is 81.4\%.

\section{Unsupervised Noisy Channel EM Method}

We next turn to a noisy-channel approach, following \citet{Knight99acomputational}.  We consider our character sequence $C = c_1 ... c_n$ as derived from a hidden (no tone) pinyin sequence $P = p_1 ... p_n$:

\vspace{0.1in}
{
    $\mbox{argmax}_\theta \ \mbox{Pr}(C) = $ \par
    $\mbox{argmax}_\theta \ \sum_P \mbox{Pr}(P) \cdot \mbox{Pr}(C | P) = $ \par
    $\mbox{argmax}_\theta \ \sum_P \mbox{Pr}(P) \cdot \prod_{i=1}^{n} \mbox{Pr}_\theta(c_i | p_i)$ 
}
\vspace{0.1in}

Pr($P$) is a fixed language model over {\em pinyin} sequences. Pr$_{\theta}$$(c | p)$ values are modified to maximize the value of the whole expression.  Examples of Pr$_{\theta}$$(c | p)$ parameters are Pr(\zh{中} $|$ {\em zhong}), which we hope to be relatively high, and Pr(\zh{很} $|$ {\em zhong}), which we hope to be zero.

\subsection{Previous Noisy-Channel}

We first faithfully re-implement \newcite{Knight99acomputational}.  They drive decipherment using a bigram Pr($P$), pruning {\em pinyin} pairs that occur few than 5~times.

Unfortunately, they do not provide their training data or code, giving only the number of character types as 2113, and the number of observed {\em pinyin} pair types as 1177 (after pruning pairs occurring fewer than 5~times).  Using our own data, we estimate their character corpus at $\sim$30,000 tokens.

We applied this re-implementation to our data.  Their {\em pinyin}-pair pruning has little effect, due to the size of our {\em pinyin} corpus (155,219 unique pairs).  We ran their expectation-maximization (EM) algorithm for 170 iterations on a character corpus of 300,000 tokens, then applied their decoding algorithm to our 6059-token test, obtaining a token pronunciation accuracy of 8.6\%.  Because this accuracy is lower than their reported 22\%, we confirmed our results with two separate implementations, and we took the best of 10~random restarts.  Increasing the character corpus size to 10m yielded a worse 5.1\% accuracy.  We conjecture that \newcite{Knight99acomputational} used more homogeneous data.

\subsection{Our Noisy-Channel}

In this work, we use a {\em pinyin}-trigram model (rather than bigram), and we apply efficiency tricks that allow us to scale to our large data.

First, we reduce our character data $C$ to a list of unique triples $C_{tri}$, recording count($c_1 c_2 c_3$) for each triple.  A sample character triple is ``\zh{的} \zh{人} \zh{口}'' (count = 43485).

Likewise, we reduce our {\em pinyin} training data to triples, sorted by unsmoothed probability Pr($p$) = normalized count($p_1 p_2 p_3$).  A sample {\em pinyin} triple is ``{\em de ren kou}'' (probability = 8$\cdot 10^{-6})$.

Our training objective now becomes:

\vspace{0.1in}

\begin{small}
$\mbox{argmax}_\theta \ \mbox{Pr}(C_{tri}) = $ \par
$\mbox{argmax}_\theta \ \prod_{c_1 c_2 c_3 \in C_{tri}} \mbox{Pr}(c_1 c_2 c_3) = $ \par
$\mbox{argmax}_\theta \ \prod_{c_1 c_2 c_3 \in C_{tri}} \\ \hspace*{0.7in} \sum_{p_1 p_2 p_3} $ \par $\mbox{Pr}(p_1 p_2 p_3) \cdot \mbox{Pr}(c_1 c_2 c_3 | p_1 p_2 p_3) = $ \par

$\mbox{argmax}_\theta \ \prod_{c_1 c_2 c_3 \in C_{tri}} $ \par 
$\hspace*{0.7in} \sum_{p_1 p_2 p_3} \mbox{Pr}(p_1 p_2 p_3) \cdot $ \par
$\hspace*{1.3in}\mbox{Pr}_\theta(c_1|p_1)\mbox{Pr}_\theta(c_2|p_2)\mbox{Pr}_\theta(c_3|p_3) $
\end{small}

\vspace{0.1in}

\begin{figure}
Given: \\ 
\hspace*{0.1in} Character triples $c_1 c_2 c_3 \in C_{tri}$, with counts\\
\hspace*{0.1in} Pinyin triples $p_1 p_2 p_3$, with probabilities \\ \\
Produce: \\
\hspace*{0.1in} Values for Pr$_{\theta}$($c | p$) \\ \\
Do: \\
\hspace*{0.1in} initialize Pr$_{\theta}$($c | p$) table (uniform, random) \\
\hspace*{0.1in} for k = 1 to max\_iterations \\
\hspace*{0.2in} initialize Pr$_k$($C_{tri}$) = 1.0 \\
\hspace*{0.2in} count($c$, $p$) = 0 \ \ (whole table)\\
\hspace*{0.2in} for each of top N character triples $c_1 c_2 c_3$ \\
\hspace*{0.3in}   sum = 0 \\
\hspace*{0.3in}     for each of top M pinyin triples $p_1 p_2 p_3$ \\
\hspace*{0.35in}       score($p_1 p_2 p_3$) = \\
\hspace*{0.4in} Pr($p_1 p_2 p_3$) Pr$_{\theta}$($c_1 | p_1$) Pr$_{\theta}$($c_2 | p_2$) Pr$_{\theta}$($c_3 | p_3$) \\
\hspace*{0.35in}       sum += score($p_1 p_2 p_3$) \\
\hspace*{0.3in} Pr$_k$($C_{tri}$) = Pr$_k$($C_{tri}$) $\cdot$ sum$^{\small \mbox{count}(c_1 c_2 c_3)}$ \\
\hspace*{0.3in} for each of top M pinyin triples $p_1 p_2 p_3$ \\
\hspace*{0.4in}   Pr($p_1 p_2 p_3 | c_1 c_2 c_3$) = score($p_1 p_2 p_3$) / sum \\
\hspace*{0.4in}   for i = 1 to 3 \\
\hspace*{0.5in}     count($c_i$, $p_i$) +=  \\
\hspace*{0.6in} Pr($p_1 p_2 p_3 | c_1 c_2 c_3$) $\cdot$ count($c_1 c_2 c_3$) \\
\hspace*{0.2in} normalize count($c$, $p$) into P$_{\theta}$($c | p$) \\
\hspace*{0.1in} return final Pr$_{\theta}$($c | p$) table
\caption{EM algorithm for revising Pr$_{\theta}$$(c | p)$ parameters ({\em pinyin}-to-character substitution probabilities) to iteratively improve the probability of observed character triples $C$.  EM guarantees Pr$_i$($C_{tri}$) $\geq$ Pr$_{i-1}$($C_{tri}$).}
\label{em}
\end{figure}

Figure~\ref{em} gives an expectation-maximization (EM) algorithm for choosing Pr$_{\theta}$$(c | p)$ to find a local maximum in Pr($C_{tri}$).  

\begin{figure}
Given: \\ 
\hspace*{0.1in} Test character string $c_1 ... c_n$ \\ 
\hspace*{0.1in} Substitution model Pr$_{\theta}$($c | p$) from EM \\ 
\hspace*{0.1in} {\em Pinyin} bigram model Pr($p_2 | p_1$) \\ \\
Produce: \\
\hspace*{0.1in} Phonetic decoding $p_1 ... p_n$ \\ \\
Do: \\
\hspace*{0.1in} Standard Viterbi algorithm \cite{viterbi} \\
\caption{Pronouncing a character sequence $c_1 ... c_n$, using a {\em pinyin} bigram model Pr($p_2 | p_1$) and EM-optimized Pr$_{\theta}$$(c | p)$ values.}
\label{viterbi}
\end{figure}

After we have obtained Pr$_{\theta}$$(c | p)$ values, we decode our 6059-character-token test sequence ($C = c_1 ... c_n$) using the standard Viterbi algorithm \cite{viterbi} Our decoding criterion is:

\vspace{0.1in}

{
    $\mbox{argmax}_P \ \mbox{Pr}(P | C) = $\par
    $\mbox{argmax}_P \ \mbox{Pr}(P) \cdot \mbox{Pr}(C | P) =$\par
    $\mbox{argmax}_P \ \mbox{Pr}(P) \cdot \prod_{i=1}^{n} \mbox{Pr}_{\theta}(c_i | p_i) \approx $\par
    $\mbox{argmax}_P \ \mbox{Pr}(P) \cdot \prod_{i=1}^{n} \mbox{Pr}_{\theta}(c_i | p_i)^3 $\par
}

\vspace{0.1in}

\noindent
where Pr($P$) is implemented with a smoothed bigram {\em pinyin} model Pr($p_2 | p_1$).  Figure~\ref{viterbi} gives the outline.  While EM only considers the top~M {\em pinyin} triples, final decoding works on entire sentences and is free to create previously-unseen {\em pinyin} trigrams.  Decoding is also free to pronounce the same character in different ways, depending on its context.  We follow the prior work in cubing channel model values.

Because different random restarts yield different accuracy results, we report ranges.  We are generally able to identify the best restart in an unsupervised way, due to the high correlation between EM's objective Pr($C_{tri}$) and test-set accuracy.

Table~\ref{emresults} shows decoding accuracy results. We achieve 71\%, substantially beating the 22\% accuracy reported by \citet{Knight99acomputational}, as well as the 8.6\% of a re-implementation applied to our data.  

\begin{table}
\centering
\small
\begin{tabular}{|r|r|r|r|} \hline 
N=M & EM iterations & Test accuracy  \\ \hline 
10k & 20 & 29 - 44 \% \\ \hline
10k & 100 & 50 \% \\ \hline
20k & 20 & 37 - 46 \% \\ \hline
20k & 100 & 58 - 62 \% \\ \hline
100k & 100 & 71 \% \\ \hline
\end{tabular}
\caption{Accuracy of noisy-channel decoding after EM training.  N is the number of unique character triples shown to EM, and M is the number of unique pinyin triples available to ``explain'' each character triple.  Accuracy ranges denote multiple random restarts.}
\label{emresults}
\end{table}

\section{Exploiting Character Components}

We next investigate whether character components can improve EM results.  Instead of storing Pr$_{\theta}$($c | p$) in a single lookup table, we compute it from five lookup tables (Pr$_1$..Pr$_5$):

\vspace{0.1in}
{
    $\mbox{Pr}_{\theta}(c | p) = $ \par
    $\hspace*{0.4in}    \lambda_1 \cdot \mbox{Pr}_1(c | p) + $ \par
    $\hspace*{0.4in}     \lambda_2 \cdot \mbox{Pr}_2(part1(c) | p) \cdot \mbox{Pr}_3(c | part1(c)) $ \par
    $\hspace*{0.4in}    \lambda_3 \cdot \mbox{Pr}_4(part2(c) | p) \cdot \mbox{Pr}_5(c | part2(c))$
}
\vspace{0.1in}

As EM establishes a tentative high value for Pr$_1$(\zh{排} $|$ {\em pai}), we hope to also create a high value for Pr$_4$(\zh{非} $|$ {\em pai}), which will encourage {\em pai} to map to other characters with component \zh{非} (such as \zh{徘}) in the following EM iterations.  

Unfortunately, while we do see compelling Pr$_4$ entries, we do not see an overall improvement in test-set accuracy from this method.

\section{Combining EM and Vector Methods}

The EM method gives 71\% accuracy, while the vector method gives 81\% accuracy.  We find that the two methods agree 47\% of time, and are 98.7\% accurate in agreement cases, so in an unsupervised way, we distill out 261 high-confidence character/pinyin mappings.

{\bf Improved EM results.}  We use the 261 high-confidence ($c$, $p$) mappings as our initial start point, by replacing each one's random initial P$_{\theta}$($c | p$) value with a 1.0 weight.  These weights bias the fractional counting in the first EM iteration.  Table~\ref{emresults2} shows that high-confidence mappings increase overall EM accuracy from 71\% to 81\%.

\begin{table}
    \centering
    \begin{small}
\begin{tabular}{|r|r|r|r|} \hline 
N=M & EM & Test accuracy & Test accuracy \\ 
& iterations & w/o hints & w/ hints \\ \hline 
20k & 20 & 37 - 46 \% & 72 - 73 \% \ \  (+27\%)\\ \hline
20k & 100 & 58 - 62 \% & 72 \% \ \  (+10\%)\\ \hline
100k & 100 &  71 \% & 81 \% \ \ (+10\%)\\ \hline
\end{tabular}
\end{small}
\caption{Improving EM results by assigning high initial weights to the 261 agreed-on mappings (``hints'') from EM and vector-based methods.  Accuracy ranges are due to multiple random restarts.  }
\label{emresults2}
\end{table}

{\bf Improved vector-based results.}  We run the same initial procedure from Section~\ref{vector}, giving us a vector space inhabited by both written words and {\em pinyin} words.  However, we modify the nearest-neighbor search that produces word/pronunciation mappings.  Our modified nearest-neighbor search takes a written word's vector and returns the nearest pronunciation vector that is consistent with the 261 high-confidence ($c$, $p$) mappings.  For example, given the word \zh{重要}, we prefer neighbors {\em dang yao} and {\em zhong yao} over {\em dang pin}, because (\zh{要},~{\em yao}) is one of the high-confidence mappings originally proposed by both EM and vector-based approaches.  We also use high-confidence mappings to improve {\em de} sequences (Section~\ref{vector}).  This combination technique is also highly effective, raising accuracy from 81\% to 89\%.





\section{Conclusion}

We implement and evaluate techniques to pronounce Chinese text in Mandarin, without the use of a pronunciation dictionary or parallel resource.  The EM method achieves a test-set accuracy of 71\%, while the vector-based method achieves 81\%.  By combining the two methods, we obtain 89\% accuracy, which significantly exceeds that of prior work.

We also demonstrate that current methods for unsupervised matching of vector spaces are sensitive to the structure of the spaces.  In the presence of one-to-many mappings between {\em pinyin} and characters, 
the mapping accuracy is severely downgraded, leaving open an opportunity to design more robust unsupervised vector mapping systems.



\bibliography{acl2019}
\bibliographystyle{acl_natbib.bst}

\end{document}